\documentclass[a4paper, 10pt, conference]{ieeeconf}      % Use this line for a4
\usepackage{FG2024}
\def\eg{{e.g.,}\@\xspace} 
\def\ie{{i.e.,}\@\xspace}

\newcommand*{\roundnum}[1]{\num[output-decimal-marker={.},
                             round-mode=places,
                             round-precision=2,
                             group-digits=false]{#1}}

\usepackage{xspace}
\usepackage{graphicx}
\usepackage{hyperref}
\usepackage{multirow, multicol}
\usepackage{siunitx}
\usepackage{booktabs}
\usepackage{subfigure}

\FGfinalcopy % *** Uncomment this line for the final submission

\IEEEoverridecommandlockouts                              % This command is only
\def\FGPaperID{****} % *** Enter the FG2024 Paper ID here

\title{\LARGE \bf
Testing the Performance of Face Recognition\\ for People with Down Syndrome
}

\author{\parbox{16cm}{\centering
    {\large C. Rathgeb$^1$, M. Ibsen$^1$, D. Hartmann$^1$, S. Hradetzky$^1$ and B. Ólafsdóttir$^2$}\\
    {\normalsize
    $^1$Hochschule Darmstadt, Darmstadt, Germany\\
    $^2$Danmarks Tekniske Universitet, Lyngby, Denmark}}
}

\usepackage{fancyhdr}
\begin{document}

\ifFGfinal
\thispagestyle{empty}
\pagestyle{empty}
\else
\author{Anonymous FG2024 submission\\ Paper ID \FGPaperID \\}
\pagestyle{plain}
\fi
\maketitle
\thispagestyle{fancy} 

%%%%%%%%%%%%%%%%%%%%%%%%%%%%%%%%%%%%%%%%%%%%%%%%%%%%%%%%%%%%%%%%%%%%%%%%%%%%%%%%
\begin{abstract}
The fairness of biometric systems, in particular facial recognition, is often analysed for larger demographic groups, e.g. female vs. male or black vs. white. In contrast to this, minority groups are commonly ignored. This paper investigates the performance of facial recognition algorithms on individuals with Down syndrome, a common chromosomal abnormality that affects approximately one in 1,000 births per year. To do so, a database of 98 individuals with Down syndrome, each represented by at least five facial images, is semi-automatically collected from YouTube. Subsequently, two facial image quality assessment algorithms and five recognition algorithms are evaluated on the newly collected database and on the public facial image databases CelebA and FRGCv2. The results show that the quality scores of facial images for individuals with Down syndrome are comparable to those of individuals without Down syndrome captured under similar conditions. Furthermore, it is observed that face recognition performance decreases significantly for individuals with Down syndrome, which is largely attributed to the increased likelihood of false matches.
\end{abstract}

%%%%%%%%%%%%%%%%%%%%%%%%%%%%%%%%%%%%%%%%%%%%%%%%%%%%%%%%%%%%%%%%%%%%%%%%%%%%%%%%
\section{Introduction}

Nowadays, systems incorporating face recognition technologies have become ubiquitous in personal, commercial, and governmental identity management applications. Recently, however, there has been a wave of public and academic concerns regarding the existence of systemic bias in face recognition. That is, algorithms have often been labelled as “racist” or “unfair” by the media, nongovernmental organizations, and researchers alike \cite{Drozdowski-BiasSurvey-TTS-2020,Rathgeb-FairnessExperts-TSM-2022}. While various methods have been proposed to improve the fairness of biometric technologies, these efforts have mainly been devoted to fairness across large demographic groups mostly related to gender and race \cite{9001031,Albiero_2020_WACV}. In contrast, only little research has been directed towards analysing the \emph{inclusiveness} of biometric technologies for minorities. 
With the increasing popularity of biometric systems, there is a growing concern that these technologies might not be inclusive towards minority groups. Many of these concerns have in common that they are often linked directly to worries about exclusion and discrimination as a result of addressing the demands of a larger user group rather than including minority groups.   

Like many other automated decision-making systems, face recognition technologies are heavily based on machine learning. These algorithms have been shown to work well on data which is similar to the data they have been trained on. However, minorities such as people with Down syndrome, are often underrepresented in the training data. With one in 1,000 births per year \cite{Morris02}, Down syndrome is the most common chromosomal abnormality in humans. The cause of this syndrome is a trisomy of all or part of chromosome 21 in all or some cells of the body \cite{Patterson09}. Certain clinical and phenotypic features are shared among affected individuals, such as congenital heart disease, cognitive impairment as well as facial appearance. Typical facial characteristics for people with Down syndrome include slanted eyes, brush marks in the iris, a round face, abnormal external ears, a flat bridge of the nose and a flattened nose, as well as a small chin \cite{Antonarakis04}, see Figure~\ref{fig:down}. While some research in the field of medical image analysis has been directed towards the detection of Down syndrome from face images \cite{Agbolade20}, a comprehensive analysis of face recognition performance for people with Down syndrome has not yet been carried out. In response, this paper makes the following contributions:

\begin{figure}[!t]
    \centering
    \includegraphics[height=2.5cm]{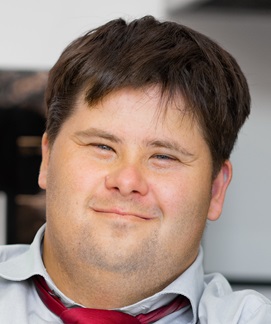}
    \includegraphics[height=2.5cm]{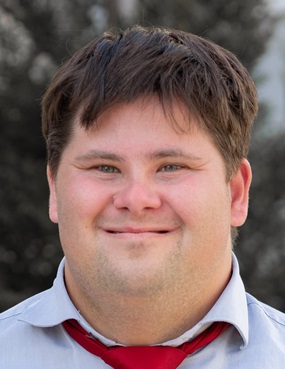}
    \hspace{0.1cm}
    \includegraphics[height=2.5cm]{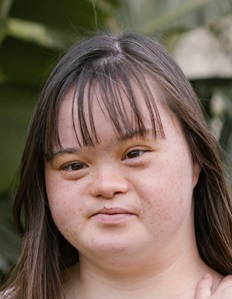}
    \includegraphics[height=2.5cm]{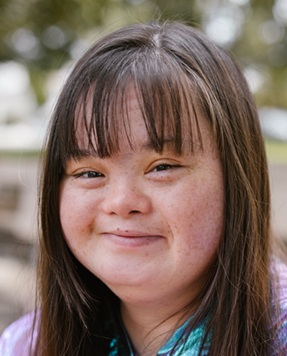}
    \caption{Example image pairs of two subjects with Down syndrome. (images licensed under CC taken from \cite{pexels}) }\label{fig:down}\vspace{-0.3cm}
\end{figure}

\begin{itemize}
    \item A collection of a database of 98 individuals with Down syndrome; in total, this database consists of 1,058 face images with at least five images per individual, which allows for a reasonable number of mated (\ie genuine) and non-mated (\ie impostor) comparisons.  
    \item A comprehensive performance analysis of face image sample quality assessment and face recognition using state-of-the-art algorithms including two open source algorithms for face quality assessment and three open source as well as two commercial-of-the-shelf (COTS) algorithms for face recognition.
    \item A comparison of the results obtained on the collected Down syndrome database with baselines on two commonly used face image databases, together with a detailed analysis of the causes of the observed performance differentials. 
\end{itemize}

This work is organised as follows: related works are briefly summarised in Sect.~\ref{sec:related}. The conducted data collection and the experimental setup are described in Sect.~\ref{sec:data} and Sect.~\ref{sec:setup}, respectively. Results are presented and discussed in Sect.~\ref{sec:results}. Finally, conclusions are drawn in Sect.~\ref{sec:conclusion}.

\section{Previous Works}\label{sec:related}

As mentioned before, some previous works have analysed the detection of Down syndrome from face images, \ie a binary classification based on a single image. While earlier methods aim at classifying features obtained from basic textural \cite{BURCIN20118690} or geometrical analysis \cite{Cornejo17}, more recent approaches are based on deep learning \cite{Gurovich19}. For a survey on proposed approaches the interested reader is referred to \cite{Agbolade20}. Databases employed in said works usually contain only a single image per subject and no identity labels are available. They only contain information whether the person is affected by Down syndrome or suffers from another gene mutation. In addition, most studies aim at an early diagnosis of Down syndrome and therefore used databases primarily contain images of children or toddlers and infants.  

With respect to the inclusiveness of biometric systems for disabled individuals, the accessibility of various security methods within a mobile banking application, including face recognition, was examined in  \cite{Acessability/Opp_biometrics}. The experiment involved the participants with multiple disabilities. All participants struggled with presenting their faces in the way required by the face recognition enrolment. Furthermore, there was a noticeable difference in error rates in the tasks between people with disabilities and the control group, which consisted of non-disabled individuals. In  \cite{physicalDisabilities_SensingSystems}, the experience of disabled people with biometric systems was analysed, including fingerprint, face, and voice recognition. More than a third of the participants stated that they had experienced face recognition to fail to recognise them or identified them as someone else. In response to such issues, the International Organization for Standardization (ISO) attempted to stipulate requirements for the inclusiveness of biometric systems \cite{iso_biometricsStandard}. In this guide, the most prevalent impairments have been listed, along with the problems these individuals might encounter and some suggestions on how these challenges might be resolved. Note that the aforementioned works mainly investigate the usability of face recognition for people with disabilities. To the best of the authors' knowledge, the performance of automated face recognition for people with disabilities (in particular Down syndrome) has not been investigated in previous works.

%Fingerprint recognition was the biometric system that caused the most issues because certain participants' disabilities made it impossible for them to maintain a particular pose or posture for the duration necessary to obtain a fingerprint of sufficient quality. Overall, there has not been much research investigating the inclusiveness of biometric technologies concerning individuals with disabilities. As a result, very little is known about this topic and how biometric systems work for this particular minority group. 

\section{Data Collection}\label{sec:data}

For the performance evaluation of face recognition algorithms, a suitable database containing face images of individuals with Down syndrome is required. Face images must fulfill the following requirements:
\begin{itemize} 
	\item The subject's face depicted in an image should be mostly frontal while common variations, \ie slight variations in pose and facial expressions, are favoured and important to investigate recognition performance.
	\item For the aspect of investigating recognition performance, unique identity labels are necessary and must therefore either already exist or be created reliably.
     \item In order to investigate the recognition performance of single individuals, several images of a subject are needed. %To ensure a certain reliability, image collections are only used if at least five images of an individual are available.
	\item For the purposes of this work, young adults and adults are primarily of interest as the primary users of facial recognition applications.
	\item Images must have a minimum resolution of 128$\times$128 pixels.
    \item A fair distribution between gender and race groups is desired.
\end{itemize}

%In the following section, the approaches taken to collect and create the required dataset are discussed. This includes the examination of already existing datasets for suitability and the possibilities to create that data collection from scratch, as well.

As already mentioned in the previous section, the existing databases do not meet the listed requirements. This motivated the compilation of a new database according to the listed criteria. To this end, the online video sharing and social media platform YouTube was used, as it has been done in various other scientific works on face recognition, \eg in \cite{6374605}. The advantage of extracting face images from videos depicting individuals with Down syndrome is that faces are traceable throughout a video. Thus identity labels can be generated for individuals and it is additionally possible to obtain several images of the same person.

To start the actual recording of the images, a link collection of videos featuring people with Down syndrome was manually created. This was done by searching for people with Down syndrome using additional keywords such as ``interview'' or ``report'' to ensure that the videos found contained appropriate facial images. The resulting link collection will be publicly released\footnote{\url{https://dasec.h-da.de/research/biometrics/hda-down-syndrome-face-database/}}. Using the Python package Selenium\footnote{\url{www.selenium.dev/selenium/docs/api/py/api.html}}, the corresponding videos are automatically opened/started and screenshots are taken in five second intervals to ensure some variance. Then, images are stored in folders according to the video title to generate corresponding ID's in a later step. Before the data are processed further, dlib \cite{King-MachineLearningToolkit-2009} is used for face detection to ensure that a face is visible. Subsequently, images that did not meet the aforementioned criteria are manually sorted out, including mainly images of individuals who are not affected by Down syndrome but are visible in the corresponding videos. %Then, (anonymized) identity labels of the individuals are manually created. It is documented in which videos the images and identities have their origin.

%To the best of our ability, we make sure that interviews or reports of affected people are collected, since frontal pictures of the people's faces can often be found here. 

%Only few works have been proposed using such a dataset, due to the private and sensitive nature of Down Syndrome subjects. Some studies have used Down Syndrome and healthy controls for binary classification, while others have used multiple genetic disorders that include Down Syndrome for multi-class classification.

%For the above mentioned reasons, we collect images with our own approaches: We use Web-Crawlers to automatically search on big online plattforms like Youtube und Instagram for digital images.

Overall, face images of 98 individuals with Down syndrome are extracted from 69 videos. The resulting database comprises a total number of 1,058 face images. It should be noted, that the number of images per person varies greatly: there are between five and 46 pictures per subject and on average about ten images. 

\section{Experimental Setup}\label{sec:setup}

The performance of commercial and open-source state-of-the-art facial analysis methods are evaluated on the collected database including face image quality assessment and recognition. Regarding quality estimation, the open-source deep learning-based algorithms FaceQnet~\cite{FaceQNet} and MagFace~\cite{MagFace} are used. In recognition experiments, AdaFace~\cite{AdaFace}, ArcFace~\cite{ArcFace}, MagFace~\cite{MagFace} and two COTS systems (COTS-1 and COTS-2) are employed. The use of the COTS algorithms raises the practical relevance of the presented benchmark. While the COTS systems are closed-source, it is assumed that these are based on deep learning, as this is the case for most of state-of-the-art methods.

Results obtained on the collected database are compared against results on two public databases: a subset of the FRGCv2 database~\cite{Phillips-OverviewFaceRecognitionGrandChallengeFRGC-CVPR-2005}, representing a constrained scenario, and the CelebA database~\cite{liu2015faceattributes}, representing a less constrained scenario. To the best of the authors' knowledge the latter two databases do not contain images of people with Down syndrome.

Biometric recognition performance is measured based on metrics defined in \cite{ISO-IEC-19795-1-Framework-210216}. Specifically,   decision thresholds at false match rates (FMRs) of 1\% and 0.1\% are reported. Note that corresponding false non-match rates (FNMRs) are not reported, since the images of the collected database generally exhibit a rather low intra-class variation since these have been extracted from single videos (as will be shown in experiments). Therefore, any conclusions drawn from mated comparisons could be misleading. On the contrary, statistics of obtained quality and recognition scores are presented. 

\section{Results}\label{sec:results}

\begin{figure}[thbp]
  \centering
  \subfigure[FaceQnet]{
    \includegraphics[width=0.62\linewidth]{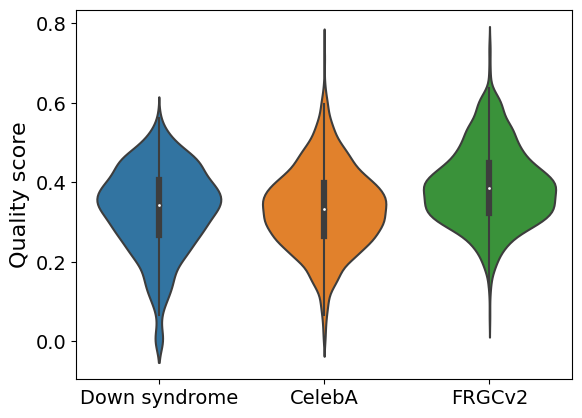}
  }
  \subfigure[MagFace]{
    \includegraphics[width=0.62\linewidth]{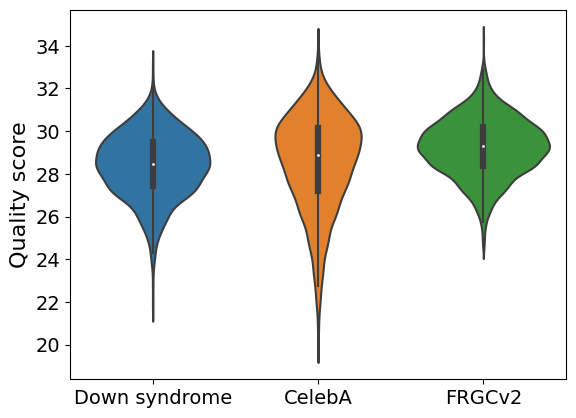}
  }
  \caption{Sample quality score distributions across datasets.}
  \label{fig:violin_quality}
\end{figure}

Average face image sample quality scores along with corresponding standards deviations for the different datasets are summarised in Table~\ref{tab:quality_metrics} (higher values indicate better sample quality). Corresponding quality score distributions are visualised in Figure~\ref{fig:violin_quality}. It can be observed that values obtained for the collected Down syndrome databases are similar to those obtained on CelebA and FRGCv2 for both of the used face image quality assessment algorithms. That is, on average the tested algorithms calculate similar quality scores for people with and without Down syndrome.

\begin{table}[htbp]
  \centering
  \caption{Sample quality statistics (average score and std. dev.) across datasets.}
    \begin{tabular}{llcc}
    \toprule
    \textbf{Algorithm} & \textbf{Database} & $\mu$ & $\sigma$ \\
    \midrule
    \multirow{3}{*}{FaceQnet} & Down syndrome & \roundnum{0.3304367548520263} & \roundnum{0.10706063724366109} \\
     & CelebA & \roundnum{0.3341268775165081} & \roundnum{0.10322143931661464} \\
     & FRGCv2 & \roundnum{0.39163023029430283} & \roundnum{0.09708959434513857} \\
    \midrule
    \multirow{3}{*}{MagFace} & Down syndrome & \roundnum{28.404557853187246} & \roundnum{1.502266994369467} \\
     & CelebA & \roundnum{28.53285789836697} & \roundnum{2.2488522339222072} \\
     & FRGCv2 & \roundnum{29.26296027055605} & \roundnum{1.3563678377224755} \\
    \bottomrule
    \end{tabular}%
  \label{tab:quality_metrics}%
\end{table}%

\begin{figure*}[thbp]
  \centering
  \subfigure[AdaFace]{
    \includegraphics[width=0.3\linewidth]{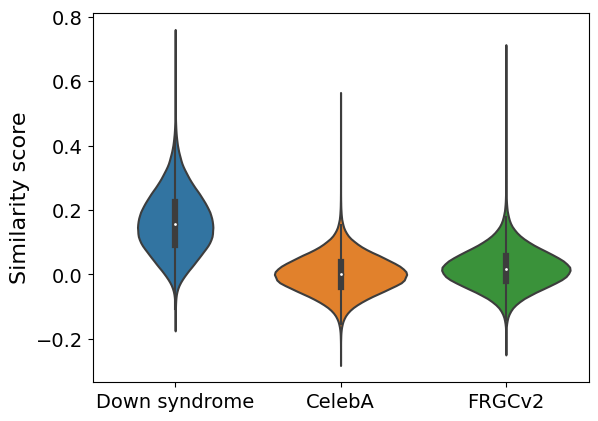}
  }
  \subfigure[ArcFace]{
    \includegraphics[width=0.3\linewidth]{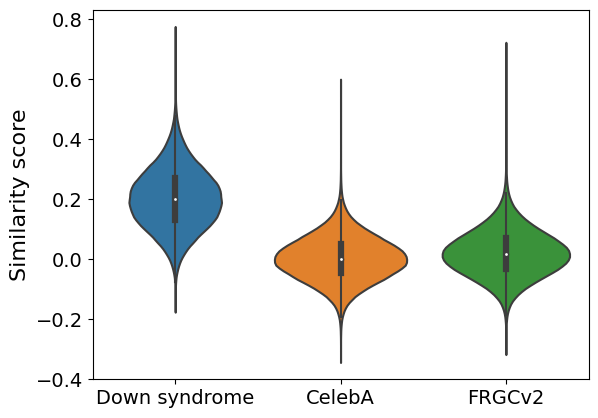}
  }
    \subfigure[MagFace]{
    \includegraphics[width=0.3\linewidth]{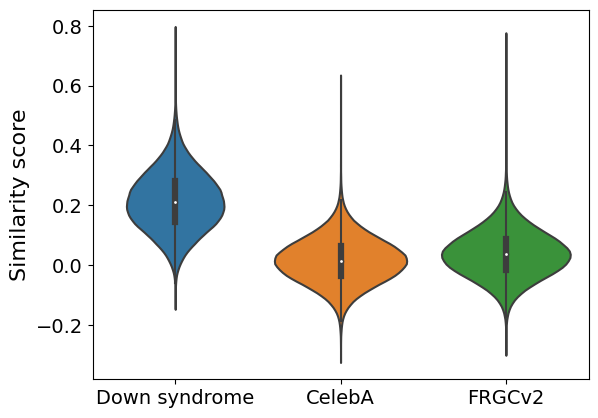}
  }
  \subfigure[COTS-1]{
    \includegraphics[width=0.3\linewidth]{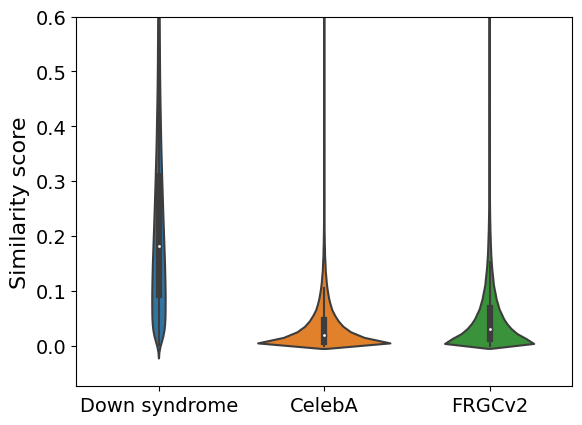}
  }
  \subfigure[COTS-2]{
    \includegraphics[width=0.3\linewidth]{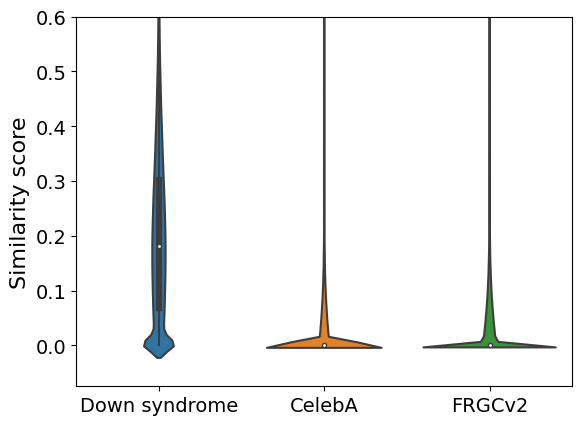}
  }
  \caption{Non-mated score distributions across datasets.}
  \label{fig:violin_recognition}
\end{figure*}

Regarding the recognition experiments, the thresholds required to achieve relevant FMRs of 0.1\% and 1\% are listed in Table~\ref{tab:fmr_thresholds}. It can be seen that for the collected Down syndrome dataset, the estimated thresholds are significantly higher compared to those of the CelebA and FRGCv2 datasets. This indicates that non-mated comparisons between subjects with Down syndrome tend to yield higher similarity scores for all of the employed face recognition systems, which is confirmed by the statistics shown in Table~\ref{tab:face_recognition_performance}. Further, non-mated score distributions of the Down syndrome dataset generally exhibit slightly higher standard deviations. A more detailed comparison between non-mated score distributions obtained by different face recognition systems is plotted in Figure~\ref{fig:violin_recognition}. 

Note that mated scores calculated for the Down syndrome dataset are generally higher compared to the ones obtained on the CelebA and FRGCv2 datasets. As mentioned earlier, this is because mated  image pairs of the Down syndrome dataset stem from frames of single videos. This means, with respect to the used datasets, the face images contained in the Down syndrome dataset have the lowest intra-class variations followed by the FRGCv2 and the CelebA datasets. 

\begin{table}[h!]
\centering
\caption{Thresholds required for relevant FMRs across datasets.}
\label{tab:fmr_thresholds}
\begin{tabular}{@{\extracolsep{2pt}}llcc@{}} \toprule 
\multicolumn{1}{c}{}  &  \multicolumn{1}{c}{}  & \multicolumn{2}{c}{\textbf{Threshold for}} \\ \cmidrule{3-4} 
\multirow{-2}{*}{\textbf{System}} & \multirow{-2}{*}{\textbf{Database}}  & \textbf{FMR$=0.1\%$}      & \textbf{FMR$=1\%$}      \\ \midrule
  \multirow{3}{*}{AdaFace} & Down syndrome & \roundnum{0.4797048139386486} & \roundnum{0.398722646626745} \\
  & CelebA & \roundnum{0.2036049628869093} & \roundnum{0.1429888507839779} \\
 & FRGCv2 & \roundnum{0.2453077009632703} & \roundnum{0.173746054297744} \\ \midrule
\multirow{3}{*}{ArcFace} &  Down syndrome & \roundnum{0.518667140897466} & \roundnum{0.44260615947504} \\
& CelebA & \roundnum{0.2478663155235757} & \roundnum{0.1784859584485511} \\
 & FRGCv2 & \roundnum{0.2919259270966606} & \roundnum{0.2104919146090601} \\ \midrule
\multirow{3}{*}{MagFace} & Down syndrome & \roundnum{0.5316096233977324} & \roundnum{0.4531233599558696} \\
& CelebA & \roundnum{0.2670922860869098} & \roundnum{0.1972305586719088} \\
 & FRGCv2 & \roundnum{0.3139103408000275} & \roundnum{0.233838403070599} \\ \midrule
\multirow{3}{*}{COTS-1} & Down syndrome & \roundnum{0.805972} & \roundnum{0.698165} \\
& CelebA & \roundnum{0.335649} & \roundnum{0.198702} \\
 & FRGCv2 & \roundnum{0.432886} & \roundnum{0.273471} \\ \midrule
\multirow{3}{*}{COTS-2} & Down syndrome & \roundnum{0.801731} & \roundnum{0.637931} \\
& CelebA & \roundnum{0.302596} & \roundnum{0.163821} \\
 & FRGCv2 & \roundnum{0.318977} & \roundnum{0.181111} \\ 
\bottomrule
\end{tabular}
\end{table}

\begin{table}[!htbp]
    \centering
    \caption{Recognition statistics for mated and non-mated distributions (average score and std. dev.) across datasets.}
        \begin{tabular}{@{\extracolsep{2pt}}llcccc@{}} \toprule 
            \multirow{2}{*}{\textbf{System}} & \multirow{2}{*}{\textbf{Database}}  & \multicolumn{2}{c}{ \textbf{Mated}} &  \multicolumn{2}{c}{ \textbf{Non-mated}} \\ \cmidrule{3-4} \cmidrule{5-6}
            & & $\mu$ & $\sigma$ & $\mu$ & $\sigma$ \\
            \midrule
            \multirow{3}{*}{AdaFace} & Down syndrome & \roundnum{0.8393529167579925} & \roundnum{0.07899516870561087} & \roundnum{0.1615505414693976} & \roundnum{0.09604626342634286} \\
             & CelebA & \roundnum{0.6081955010400858} & \roundnum{0.17766250969321293} & \roundnum{0.0010899030185480031} & \roundnum{0.058136957860781706} \\
             & FRGCv2 & \roundnum{0.7114530933017618} & \roundnum{0.07080863619555687} & \roundnum{0.01916373412955203} & \roundnum{0.060660374905408765} \\ \midrule 
            \multirow{3}{*}{ArcFace} & Down syndrome & \roundnum{0.8457458617275432} & \roundnum{0.07731615477634385} & \roundnum{0.20171546639461752} & \roundnum{0.10007804260038933} \\
             & CelebA & \roundnum{0.6357083467709858} & \roundnum{0.17868359679296175} & \roundnum{0.0014876317453078759} & \roundnum{0.07321525963542193} \\
             & FRGCv2 & \roundnum{0.7494351853542607} & \roundnum{0.062257176499803425} & \roundnum{0.019064061162370225} & \roundnum{0.07678707051593299} \\ \midrule 
            \multirow{3}{*}{MagFace} & Down syndrome & \roundnum{0.8527451269772436} & \roundnum{0.07503483158015718} & \roundnum{0.21388330584716786} & \roundnum{0.09866691552894406} \\
             & CelebA & \roundnum{0.6565028290018362} & \roundnum{0.17691690762614956} & \roundnum{0.015311946302785283} & \roundnum{0.07571365380604597} \\
             & FRGCv2 & \roundnum{0.763278848493316} & \roundnum{0.060421685573080024} & \roundnum{0.03762756899315124} & \roundnum{0.07872219073785315} \\ \midrule 
            \multirow{3}{*}{COTS-1} & Down syndrome & \roundnum{0.9800150747224595} & \roundnum{0.020958487439351338} & \roundnum{0.22049605039789097} & \roundnum{0.16420083658925347} \\
             & CelebA & \roundnum{0.8700466124299722} & \roundnum{0.2083412565369612} & \roundnum{0.034358238958111235} & \roundnum{0.04264034031186373} \\
             & FRGCv2 & \roundnum{0.963990915709598} & \roundnum{0.014658762851470887} & \roundnum{0.050136145569260125} & \roundnum{0.058176343230257076} \\ \midrule 
            \multirow{3}{*}{COTS-2} & Down syndrome & \roundnum{0.991502313692001} & \roundnum{0.03865523149176314} & \roundnum{0.19989949610061972} & \roundnum{0.15930162212044122} \\
             & CelebA & \roundnum{0.9056989563291792} & \roundnum{0.2159248333036219} & \roundnum{0.008863744591991511} & \roundnum{0.03190002502146565} \\
             & FRGCv2 & \roundnum{0.9923253160377357} & \roundnum{0.004981967577903247} & \roundnum{0.011866174670023584} & \roundnum{0.03656347454659115} \\ \bottomrule 
        \end{tabular}
    \label{tab:face_recognition_performance}
\end{table}
\newpage
In summary, the following observations have been made for the tested algorithms:
\begin{itemize}
    \item Face image sample quality scores for people with Down syndrome are in the same range as those obtained for subjects without Down syndrome. This means, with respect to quality assessment, no disadvantage is observable for people with Down syndrome.
    \item Albeit being extracted from single videos (which equals a single acquisition session), face images of the collected Down syndrome face database yield high similarity scores for mated comparisons. This indicates that subjects with Down syndrome are not expected to experience higher rejection rates (in terms of FNMR) when using face recognition technologies.
    \item Similarity scores resulting from non-mated comparisons are significantly higher for people with Down syndrome. As mentioned earlier, subjects with Down syndrome share certain facial characteristics, which appear to cause this effect. As a result, the probability of false accepts (in terms of FMR) increases which impairs the security of face recognition for individuals with Down syndrome. In case decision thresholds are adjusted accordingly, higher false rejection rates are expected.
\end{itemize}

\section{Conclusion}\label{sec:conclusion}

With the evolution of biometrics, ethical concerns have emerged with respect to the fairness or inclusiveness of these technologies. While many research efforts have been devoted towards investigating race and gender biases, little to no efforts have been put into studying the inclusiveness of biometric systems for disabled individuals.

This work provided a preliminary study on the performance of facial recognition technologies for people with Down syndrome. While the results showed that face image quality assessment algorithms are applicable to this minority group, this was not the case for automated face recognition. Across five different face recognition algorithms it was also found that individuals with Down syndrome suffer from higher probabilities of false matches. To mitigate the observed bias, face images of subjects with Down syndrome should be included in training databases of face recognition technologies. In this context, the database collected in this work can serve as a starting point for future research. 
\newpage

\section{Acknowledgements}

This research work has been partially funded by the Hessian Ministry of the Interior and Sport in the course of the Bio4ensics project, by German Federal Ministry of Education and Research and the Hessian Ministry of Higher Education, Research, Science and the Arts within their joint support of the National Research Center for Applied Cybersecurity, ATHENE.

{\small
\bibliographystyle{ieee}
\bibliography{references}

\begin{thebibliography}{10}\itemsep=-1pt

\bibitem{Agbolade20}
O.~Agbolade, A.~Nazri, R.~Yaakob, A.~A. Ghani, and Y.~K. Cheah.
\newblock Down syndrome face recognition: A review.
\newblock {\em Symmetry}, 12(7), 2020.

\bibitem{Albiero_2020_WACV}
V.~Albiero, K.~K.S., K.~Vangara, K.~Zhang, M.~C. King, and K.~W. Bowyer.
\newblock Analysis of gender inequality in face recognition accuracy.
\newblock In {\em IEEE/CVF Winter Conference on Applications of Computer Vision
  (WACV) Workshops}, 2020.

\bibitem{Antonarakis04}
S.~E. Antonarakis, R.~Lyle, E.~T. Dermitzakis, A.~Reymond, and S.~Deutsch.
\newblock Chromosome 21 and down syndrome: From genomics to pathophysiology.
\newblock {\em Nature Reviews Genetics}, 5:725–--738, 2004.

\bibitem{Acessability/Opp_biometrics}
R.~Blanco-Gonzalo, C.~Lunerti, R.~Sanchez-Reillo, and R.~Guest.
\newblock Biometrics: Accessibility challenge or opportunity?
\newblock {\em PloS one}, 13(3):e0194111, 2018.

\bibitem{BURCIN20118690}
K.~Burçin and N.~V. Vasif.
\newblock Down syndrome recognition using local binary patterns and statistical
  evaluation of the system.
\newblock {\em Expert Systems with Applications}, 38(7):8690--8695, 2011.

\bibitem{Cornejo17}
J.~Y.~R. Cornejo, H.~Pedrini, and F.~D. L. d. S.~N. A.~Machado-Lima.
\newblock Down syndrome detection based on facial features using a geometric
  descriptor.
\newblock {\em Journal of Medical Imaging}, 4, 2017.

\bibitem{6374605}
A.~Dantcheva, C.~Chen, and A.~Ross.
\newblock Can facial cosmetics affect the matching accuracy of face recognition
  systems?
\newblock In {\em International Conference on Biometrics: Theory, Applications
  and Systems (BTAS)}, pages 391--398, 2012.

\bibitem{ArcFace}
J.~Deng, J.~Guo, J.~Yang, N.~Xue, I.~Kotsia, and S.~Zafeiriou.
\newblock Arcface: Additive angular margin loss for deep face recognition.
\newblock {\em IEEE Transactions on Pattern Analysis and Machine Intelligence},
  44(10):5962--5979, 2022.

\bibitem{Drozdowski-BiasSurvey-TTS-2020}
P.~Drozdowski, C.~Rathgeb, A.~Dantcheva, N.~Damer, and C.~Busch.
\newblock Demographic bias in biometrics: A survey on an emerging challenge.
\newblock {\em Trans. on Technology and Society ({TTS})}, 1(2):89--103, 2020.

\bibitem{iso_biometricsStandard}
I.~O. for Standardisation.
\newblock Information technology — biometric performance testing and
  reporting) -- {Part} 1: {Principles and framework}.
\newblock Standard ISO/IEC 19795-1:2021(E), 2021.

\bibitem{Gurovich19}
Y.~Gurovich, Y.~Hanani, O.~Bar, N.~F. Guy~Nadav, D.~Gelbman, L.~Basel-Salmon,
  P.~M. Krawitz, S.~B. Kamphausen, M.~Zenker, and L.~M. B. A. K.~W. Gripp.
\newblock {\em Nature Medicine}, page 60–64, 2019.

\bibitem{FaceQNet}
J.~Hernandez-Ortega, J.~Galbally, J.~Fierrez, R.~Haraksim, and L.~Beslay.
\newblock {FaceQnet}: Quality assessment for face recognition based on deep
  learning.
\newblock In {\em International Conference on Biometrics (ICB)}, 2019.

\bibitem{ISO-IEC-19795-1-Framework-210216}
{ISO/IEC JTC1 SC37 Biometrics}.
\newblock {\em {ISO/IEC} 19795-1:2021. Information Technology -- Biometric
  Performance Testing and Reporting -- Part~1: Principles and Framework}, 2021.

\bibitem{physicalDisabilities_SensingSystems}
S.~K. Kane, A.~Guo, and M.~R. Morris.
\newblock Sense and accessibility: Understanding people with physical
  disabilities’ experiences with sensing systems.
\newblock In {\em 2nd Int'ernationa'l ACM SIGACCESS Conference on Computers and
  Accessibility}, pages 1--14, 2020.

\bibitem{AdaFace}
M.~Kim, A.~K. Jain, and X.~Liu.
\newblock {AdaFace}: Quality adaptive margin for face recognition.
\newblock In {\em IEEE/CVF Conference on Computer Vision and Pattern
  Recognition (CVPR)}, 2022.

\bibitem{King-MachineLearningToolkit-2009}
D.~King.
\newblock Dlib-ml: A machine learning toolkit.
\newblock {\em Journal of Machine Learning Research}, 2009.

\bibitem{9001031}
K.~S. Krishnapriya, V.~Albiero, K.~Vangara, M.~C. King, and K.~W. Bowyer.
\newblock Issues related to face recognition accuracy varying based on race and
  skin tone.
\newblock {\em IEEE Transactions on Technology and Society}, 1(1):8--20, 2020.

\bibitem{liu2015faceattributes}
Z.~Liu, P.~Luo, X.~Wang, and X.~Tang.
\newblock Deep learning face attributes in the wild.
\newblock In {\em Proceedings of International Conference on Computer Vision
  (ICCV)}, 2015.

\bibitem{MagFace}
Q.~Meng, S.~Zhao, Z.~Huang, and F.~Zhou.
\newblock {MagFace}: A universal representation for face recognition and
  quality assessment.
\newblock In {\em IEEE/CVF Conference on Computer Vision and Pattern
  Recognition (CVPR)}, 2021.

\bibitem{Morris02}
J.~K. Morris, D.~E. Mutton, and E.~Alberman.
\newblock Revised estimates of the maternal age specific live birth prevalence
  of {D}own’s syndrome.
\newblock {\em Journal of Medical Screening}, 9(1):2--6, 2020.

\bibitem{Patterson09}
D.~Patterson.
\newblock Molecular genetic analysis of {Down} syndrome.
\newblock {\em Human Genetics}, 126(1):195--214, 2009.

\bibitem{pexels}
{Pexels Free Stock Photos}.
\newblock https://www.pexels.com, 2024.

\bibitem{Phillips-OverviewFaceRecognitionGrandChallengeFRGC-CVPR-2005}
J.~Phillips, P.~Flynn, T.~Scruggs, K.~Bowyer, J.~Chang, et~al.
\newblock Overview of the {{Face Recognition Grand Challenge}}.
\newblock In {\em Conf. on {{Computer Vision}} and {{Pattern Recognition}}
  ({{CVPR}})}, pages 947--954. IEEE, June 2005.

\bibitem{Rathgeb-FairnessExperts-TSM-2022}
C.~Rathgeb, P.~Drozdowski, D.~C. Frings, N.~Damer, and C.~Busch.
\newblock Demographic fairness in biometric systems: What do the experts say?
\newblock {\em {IEEE} Technology and Society Magazine}, 41:71--82, 2022.

\end{thebibliography}
}

\end{document}